\documentclass[10pt,twocolumn,letterpaper]{article}
\usepackage[pagenumbers]{wacv} 

\usepackage{amsmath}
\usepackage{amssymb}
\usepackage{booktabs}
\usepackage{adjustbox}
\usepackage[hidelinks]{hyperref}
\usepackage{balance}
\usepackage[table]{xcolor}

\begin{document}

\title{SilVar-Med: A Speech-Driven Visual Language Model for Explainable Abnormality Detection in Medical Imaging}


\author{Tan-Hanh Pham$^{1, 2, *, \dag}$  $\quad $Chris Ngo$^2$ $\quad $Trong-Duong Bui$^3$ \\Minh Luu Quang$^4$ $\quad $Tan-Huong Pham$^5$   $\quad $Truong-Son Hy$^{6, *}$\\
\hphantom{text}\\ 
{\fontsize{11}{11}\selectfont$^1$Florida Institute of Technology, USA, $^2$Knovel Engineering Lab, Singapore, }
\\{\fontsize{11}{11}\selectfont$^3$Vietnam Military Medical University, $^4$108 Military Central Hospital, Vietnam,}
\\{\fontsize{11}{11}\selectfont$^5$Can Tho University of Medicine and Pharmacy, Vietnam, $^6$University of Alabama at Birmingham, USA}
\hphantom{text}
}

\maketitle

\begin{abstract}

{\fontsize{10}{10}\selectfont Medical Visual Language Models have shown great potential in various healthcare applications, including medical image captioning and diagnostic assistance. However, most existing models rely on text-based instructions, limiting their usability in real-world clinical environments especially in scenarios such as surgery, text-based interaction is often impractical for physicians. In addition, current medical image analysis models typically lack comprehensive reasoning behind their predictions, which reduces their reliability for clinical decision-making. Given that medical diagnosis errors can have life-changing consequences, there is a critical need for interpretable and rational medical assistance. To address these challenges, we introduce an end-to-end speech-driven medical VLM, SilVar-Med, a multimodal medical image assistant that integrates speech interaction with VLMs, pioneering the task of voice-based communication for medical image analysis. In addition, we focus on the interpretation of the reasoning behind each prediction of medical abnormalities with a proposed reasoning dataset. Through extensive experiments, \textbf{we demonstrate a proof-of-concept study for reasoning-driven medical image interpretation with end-to-end speech interaction}. We believe this work will advance the field of medical AI by fostering more transparent, interactive, and clinically viable diagnostic support systems. Our code and dataset are publicly available at \href{https://github.com/Hanhpt23/SilVarMed.git}{SiVar-Med}.}
\end{abstract}

\def\thefootnote{*}\footnotetext{Corresponding authors: hanhpt.phamtan@gmail.com, thy@uab.edu}\def\thefootnote{\arabic{footnote}}
\def\thefootnote{$\dag$}\footnotetext{Project lead}\def\thefootnote{\arabic{footnote}}


\section{Introduction}
\label{sec.intro}
Recently, advancements in Visual Language Models (VLMs) have demonstrated the potential of Large Language Models (LLMs) to process both images and text at the same time \cite{li2023blip, ranasinghe2023language, alayrac2022flamingo, awadalla2023openflamingo, NEURIPS2023_6dcf277e}. In the medical domain, VLMs have gained increasing attention for their ability to facilitate intuitive human-machine interactions such as MedBLIP \cite{chen2024medblip}, Med-flamingo~\cite{moor2023med}, Llava-Med \cite{li2024llava}, improving clinical decision-making and diagnostic assistance. These models are particularly valuable for medical imaging analysis, where they can process complex radiological images — such as X-ray \cite{le2024litegpt, hyland2023maira}, MRI, and CT scans \cite{xie2024medtrinity} — and generate meaningful textual descriptions. By leveraging deep learning techniques, VLMs can assist professionals in interpreting medical images, identifying abnormalities, and supporting diagnostic workflows.

Despite these advancements, most existing medical VLMs remain limited to text-based interactions, which may not be optimal in time-sensitive clinical settings or for visually impaired users. While some proprietary VLMs, such as GPT-4o \cite{openai2024gpt4} and Gemini \cite{team2023gemini}, support speech-driven interactions, they are not open-source, restricting fine-tuning for downstream tasks. Recently, SilVar \cite{pham2024SilVar}, a speech-driven multimodal model for reasoning-based visual question answering and object localization, has emerged as a pioneering effort in the field. Despite its potential applications in the medical domain, speech-based medical instruction for VLMs remains underexplored in open-source research, and existing models lack the capability to process and reason through spoken queries effectively.

Furthermore, while there are some several benchmarks, have been introduced to evaluate the performance of medical VLMs such as MultiMedEval \cite{royer2024multimedeval}, MultiMedQA \cite{singhal2022large}, OmniMedVQA \cite{hu2024omnimedvqa}, existing evaluation methods primarily focus on image captioning tasks and are limited in assessing the reasoning behind predictions. In addition, commonly used datasets such as SLAKE \cite{liu2021slake}, VQA-Med \cite{ImageCLEFVQA-Med2019}, VQA-RAD \cite{lau2018dataset}, and PathVQA \cite{he2020pathvqa} primarily evaluate models using text-based instructions with short-answer responses, often without requiring deeper reasoning or justification. To address this limitation, LLaVA-Med \cite{li2024llava} introduced a medical chat assistant capable of answering open-ended research questions, but its functionality is restricted to image-text inputs and limited by the number of supported medical image modalities. OmniMedVQA \cite{hu2024omnimedvqa}, on the other hand, aggregates multiple available datasets to create a larger benchmark for multiple-choice question-answering tasks and utilizes LLMs as judges for evaluation. However, both approaches lack a structured framework for evaluating reasoning abilities in medical VLMs, particularly in the context of abnormality detection.

To bridge this gap, we propose SilVar-Med, an end-to-end speech-instructed medical VLM that enables users to interact with the model verbally. Our approach not only introduces speech-driven interaction but also concentrates in reasoning abnormality detection by incorporating structured reasoning into predictions. To this end, we introduce a demonstrated dataset, designed for reasoning abnormality detection through speech instructions. Additionally, we propose a novel evaluation metric that leverages LLMs as judges to assess the reasoning capabilities of medical VLMs. Our contribution is summarized as follows:
\begin{itemize} 
    \item We propose SilVar-Med, a speech-driven medical VLM that enables intuitive human-machine interaction in healthcare. 
    \item We focus on investigating the model's reasoning abilities behind abnormality detection, addressing the limitations of predictions without explanations or short predictions. 
    \item We introduce a dataset for speech-instructed medical abnormality detection, enhancing multimodal learning in medical AI. 
    \item We propose a comprehensive reasoning evaluation metric together with LLMs as judges for medical VLMs. 
\end{itemize}


\section{Related Work}
\label{sec.relatedwork}

\subsection{Medical Vision Language Models}
Over the past five years, there has been a rapid development of LLMs and VLMs such as Gemini and GPT-4 \cite{brown2020language, achiam2023gpt}, alongside the emergence of open-source models like the Llama family \cite{touvron2023Llama, touvron2023Llama2, dubey2024Llama}, Mistral family \cite{jiang2023mistral}, Qwen family \cite{bai2023qwen, yang2024qwen2}, and Vicuna \cite{zheng2023judging}. These models have significantly advanced natural language understanding, but their capabilities have been further extended by VLMs, which integrate visual and textual modalities \cite{li2023blip}. VLMs enable models to process both images and text, enhancing applications such as visual question answering (VQA), medical image interpretation, and image captioning. There are many VLMs including Flamingo \cite{alayrac2022flamingo}, BLIP \cite{liu2024improved}, MiniGPT-v2 \cite{chen2023minigpt}, MiniGPT-4 \cite{zhu2023minigpt}, LLaVA \cite{NEURIPS2023_6dcf277e}, and InternVL \cite{chen2024internvl}, which have demonstrated remarkable progress in general-domain visual-language tasks.

Inspired by these advancements, researchers have developed domain-specific VLMs for medical applications \cite{zhang2023biomedclip, chen2024medblip, windsor2023visionlanguage}. One of the pioneering studies in this field is Med-Flamingo \cite{moor2023med}, which extends Flamingo to the medical domain by pretraining on multimodal knowledge sources spanning various medical disciplines. Similarly, LLaVA-Med \cite{li2024llava} filters image-text pairs from PMC-15M \cite{zhang2023biomedclip} to train a biomedical-specialized VLM leveraging LLaVA-pretrained parameters. In addition to medical image report generation and medical image captioning models \cite{you2023cxr}, MiniGPT-Med \cite{alkhaldi2024minigpt} and Lite-GPT \cite{le2024litegpt} extend MiniGPTs \cite{chen2023minigpt, zhu2023minigpt} to generate bounding boxes along with predictions, enabling localized abnormality detection. Furthermore, Merlin \cite{blankemeier2024merlin} is one of the pioneering models for 3D VLMs, capable of processing 3D medical images alongside their corresponding textual radiology reports, along with RadFM \cite{wu2023towards}. Other notable studies, such as PubMedCLIP \cite{eslami2023pubmedclip}, BiomedCLIP \cite{zhang2023biomedclip}, and BiomedGPT \cite{zhang2023biomedgpt}, have also contributed to the adaptation of general-domain VLMs for medical applications.

However, most of these work underexplored the reasoning behind prediction and concentrate on short answer generation or multiple choice, reducing their reliability for clinical decision-making. In addition, medical VLMs remain limited to image-text interactions, which may not be convenient in scenarios where text input is unavailable or impractical. For example, in surgical environments, speech-based interactions could be more effective, as verbal communication is often preferred over manual text input.

\subsection{Medical Datasets and Benchmarks}
In addition to model development, researchers have made efforts to create medical VQA datasets to support the ongoing advancements in the field. Several fundamental and widely used medical VQA datasets have been developed, including SLAKE \cite{liu2021slake}, VQA-RAD \cite{lau2018dataset}, PathVQA \cite{he2020pathvqa}, VQA-Med (2018–2021) \cite{ImageCLEFVQA-Med2019}, and PubMedQA \cite{jin2019pubmedqa}, EHRXQA \cite{bae2023ehrxqa}. However, these datasets are often limited in size or lack diversity in medical imaging modalities. To address these limitations, recent studies have attempted to scale dataset size using GPT-assisted models and prompting techniques. For example, models such as LLaVA-Med \cite{li2024llava} and MedTrinity \cite{xie2024medtrinity} have leveraged large-scale dataset generation through synthetic data augmentation. Furthermore, OmniMedVQA \cite{hu2024omnimedvqa} combines both published and restricted datasets to provide a diverse and large-scale medical VQA benchmark, primarily focusing on multiple-choice questions. In addition, PMC-VQA \cite{zhang2023pmc} was generated using self-instruction on PMC-OA \cite{lin2023pmc}, offering a comprehensive dataset for biomedical VQA tasks.

Despite these efforts, current medical VLMs still struggle with reasoning-based predictions, resulting in medical VQA models excelling at image captioning tasks but lacking structured reasoning mechanisms to justify their outputs. Moreover, existing evaluation methods primarily focus on text similarity and alignment metrics (n-grams) such as accuracy, BLEU, and ROUGE, without adequately assessing the depth of reasoning in model predictions. These metrics may also fail to capture the semantic quality and logical coherence of the model’s reasoning process.

To address these challenges, in this work, we propose SilVar-Med, an end-to-end speech-instructed medical VLM that enhances multimodal interactions and supports structured reasoning for abnormality detection. In addition, we focus on reasoning-based abnormality detection that improves model transparency and decision-making reliability. To this end, we introduce a demonstration dataset for reasoning-based abnormality detection. In term of evaluation, we propose using LLMs as a judge framework with a focus on reasoning responses.

\section{Data Processing}
\label{sec.data}

\subsection{Reasoning Abnormality Dataset}
\label{subsec.abnormal_detection_dataset}

To achieve our study's objective - developing a medical assistant that understands medical images and enables users to interact with it through voice queries - we created a demo dataset addressing two key challenges: (1) understanding the reasoning behind each abnormality detection and (2) enabling voice-based instructions or queries. Particularly, we focus on abdominal and thoracic  abnormalities detection across three imaging modalities: MRI, CT, and X-ray. Our dataset includes abnormalities in six organs, including heart, liver, kidney, lung, spleen, because they align well with the expertise of the physicians on our team.

Recognizing the ability of large language models (LLMs) to effectively learn from visual features in images and their corresponding reasoning descriptions, we intentionally created a small, specialized dataset tailored for our downstream task. We manually selected abnormal samples from the SLAKE dataset \cite{liu2021slake} and then constructed our explainable abnormality detection dataset. Initially, three medical image analysis specialists from our team manually annotated the dataset. However, after annotating a substantial number of samples, we found that we could leverage GPT-4o with Chain-of-Thought (CoT) prompting, combined with our medical expertise, to enhance quality and efficiency. Eventually, the annotation process is described into four steps as follows:

\begin{itemize}
    \item \textbf{Data Selection}: Identifying and verifying abnormal samples extracted from the SLAKE dataset. Following the initial annotation, we select samples that exhibit abnormalities and confirmed disease diagnoses.
    
    \item \textbf{Annotation \& Labeling}: After selecting the desired samples, we identify bounding boxes and label abnormalities directly on the images. Then, we leverage GPT-4o and CoT prompting to generate preliminary annotations as mentioned above. The purpose of this step is to leverage the knowledge of GPT-4o to assist in labeling.
    
    \item  \textbf{Specialists Validation}: Using the original labels and preliminary annotations, our team of three medical image analysis specialists manually reviewed and relabeled the data to generate high-quality annotations, ensuring the correctness of all labels (866 samples).
    \item \textbf{Synthetic Voice Generation}: We normalize the text in the questions to ensure that it generates smooth, natural speech outputs, enhancing the clarity and coherence of the spoken content before using Google Cloud APIs to generate synthetic voice. Regarding the naturalness of the sound, we believe it is more related to sentiment and emotional recognition, which is beyond the scope of this study. Therefore, we assume that the nuance and tone of the synthetic voice do not significantly affect our task, abnormality prediction.
\end{itemize}
Finally, our dataset has the following imaging modality distribution: MRI (22.4\%, 194 samples), CT (16.5\%, 143 samples), and X-ray (61.1\%, 529 samples). The dataset is split into training and testing sets: train (162 MRI, 122 CT, 432 X-ray) and test (32 MRI, 21 CT, 97 X-ray).

\small
The dataset is designed to support reasoning-based abnormality detection, with questions structured to extract meaningful insights about abnormalities in an organ. 

\textbf{Example questions include}:
\begin{itemize}
\item What diseases are present in the image?
\item Does the {specific organ} appear normal or abnormal?
\item Is the {specific organ} healthy?
\item Are there any abnormalities in this image?
\item Which organ is abnormal: \{specific organ\} or \{specific organ\}?
\end{itemize}
\textbf{For the answer structure, we ensure that}:
\begin{itemize}
\item The first sentence directly answers the question.
\item The subsequent sentences provide reasoning, explaining the signs of abnormality.
\end{itemize}

\begin{table}[ht]
\centering
\begin{tabular}{>{\arraybackslash}m{5.5cm} >{\centering\arraybackslash}m{2cm}}
\toprule
\rowcolor{gray!10} \multicolumn{2}{l}{\textbf{Sample 1}: No abnormality detection.} \\
\textbf{Question}: Is the lung healthy?\vspace{0.2cm}

\textbf{Answer}: The lung appears healthy based on the provided imaging. There are no visible signs of abnormalities such as lesions, fluid accumulation, or structural changes. The lung fields are clear, indicating normal function and absence of disease. & 
\includegraphics[width=2.3cm]{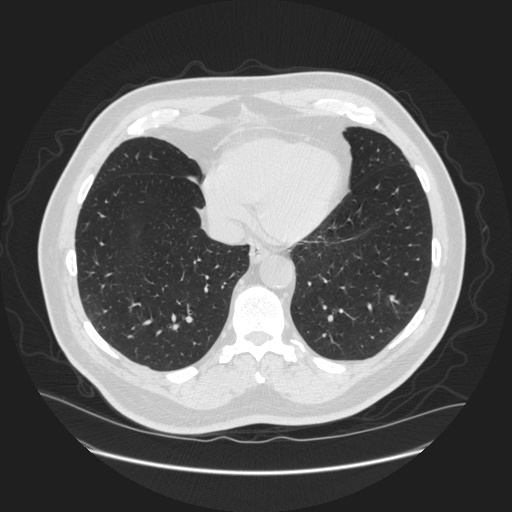} \\ 
\midrule
\rowcolor{gray!10} \multicolumn{2}{l}{\textbf{Sample 2}: Abnormality detection.} \\
\textbf{Question}: What diseases are included in the picture?\vspace{0.2cm}

\textbf{Answer}: The image indicates the presence of lung cancer, characterized by abnormal growths in the lung tissue. The CT scan shows potential lesions or nodules that may suggest malignancy.& 
\includegraphics[width=2.3cm]{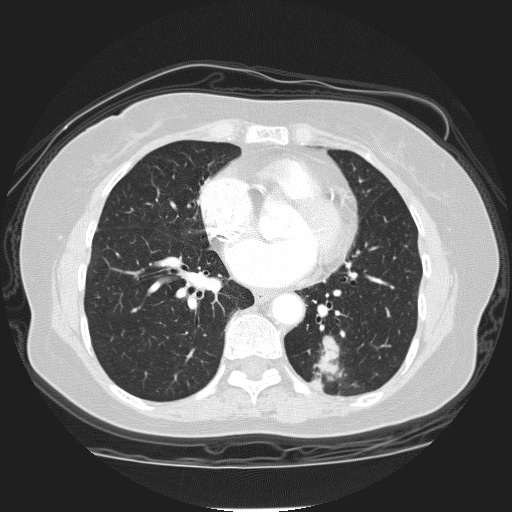} \\ 
\bottomrule
\end{tabular}
\caption{Examples of reasoning abnormal detection.}
\label{tab:chatbot}
\end{table}

\subsection{Data Collection}

\begin{table}[h]
    \centering
    \small
    \begin{tabular}{lccc}
        \toprule
        Dataset & Train & Validation & Test \\
        \midrule
        VQA-RAD \cite{lau2018dataset} & 1.7k & None & 451 \\
        SLAKE (English)\cite{liu2021slake} & 4.9k & 1k & 1k \\
        VQA-Med 2019 \cite{ImageCLEFVQA-Med2019} & 12.7k & 2k & 500 \\
        SilVar-Med (ours) & 716 & - & 150 \\
        \bottomrule
    \end{tabular}
    \caption{Summary of Medical VQA Datasets.}
    \label{tab:medical_vqa}
\end{table}

Recognize the size limitation of our reasoning dataset, we leverage some published datasets including SLAKE \cite{liu2021slake}, VQA-RAD \cite{lau2018dataset}, VQA-Med 2019 \cite{ImageCLEFVQA-Med2019} to enhance its learning knowledge. For these dataset, we follow the synthetic voice generation pipeline as mentioned in \ref{subsec.abnormal_detection_dataset}. For the SLAKE dataset, we use only english samples.

\section{Reasoning Abnormality Detection}
\subsection{Speech-Driven Medical VLM}

\begin{figure*}[ht]
    \centering
    \includegraphics[width=0.8\linewidth]{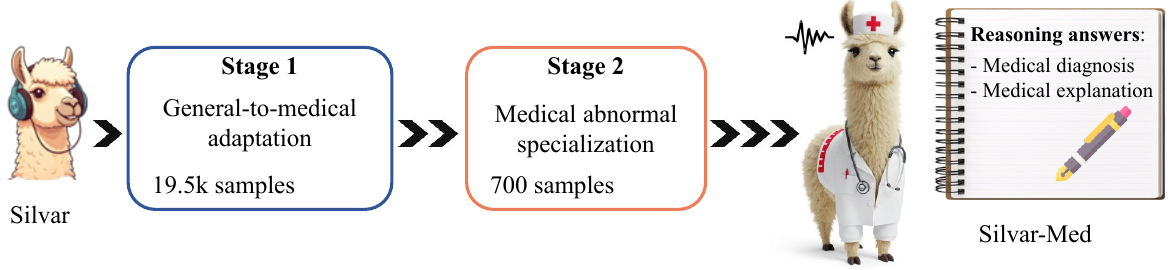}
    \caption{SilVar-Med training pipline.}
    \label{fig:SilVar-med}
\end{figure*}

Our SilVar-Med is a multimodal model built upon to SilVar \cite{pham2024SilVar}, for medical abnormality detection by integrating speech and image inputs. Unlike traditional medical visual language models that rely on text-based instructions, SilVar-Med introduces an end-to-end speech-driven approach, making it more suitable for real-world clinical environments where text interaction is impractical, such as in surgical settings. Inherit from the flexibility of the modules in the SilVar model, we designed the model with three key components: an audio encoder that extracts speech features, a visual encoder that processes medical images, and a large language model that fuses multimodal inputs to generate reasoned text responses for abnormality detection. In addition, we modified the vision encoder with PubMedCLIP \cite{eslami2023pubmedclip}, and the language model with Deepseek R1 (Distill-8B-Llama) \cite{deepseekR1}, a rising star for reasoning response. By combining speech and vision-based reasoning, SilVar-Med enhances interpretability in medical imaging, providing a more interactive and transparent diagnostic support system.

\subsection{Training Pipeline}
\label{sec.train_pipeline}

The training of SilVar-Med follows a two-stage process, as shown in \ref{fig:SilVar-med}. In the first stage, general-to-medical adaptation, we train the Whisper \cite{radford2023robust} model with a speech-to-text task in the medical domain to ensure it effectively extracts meaningful features from spoken instructions. Once trained, the Whisper encoder is integrated into SilVar-Med, where it works alongside the medical visual encoder and language model to process multimodal inputs. After that, we train the SilVar-Med with 19.5k English medical VQA samples as we mentioned in \ref{tab:medical_vqa}. In the second stage, we continue trainning the model with our dataset, specializing in medical abnormality detection and reasoning-based medical image interpretation.

In terms of training configuration, we conducted experiments with the Tiny and Small Whisper models for 20 epochs using a batch size of 8. For SilVar-Med, we employ a weight decay of 0.05 and train the model for 20 epochs, with each epoch consisting of 177 iterations. The learning rate is set to 1e-5 and remains constant throughout training, with both the minimum and warmup learning rates also set to 1e-5.  Each training batch consists of four samples, and the training utilizes two workers to optimize computational efficiency. This structured training approach ensures that SilVar-Med effectively learns from diverse medical datasets and refines its reasoning capabilities through targeted fine-tuning.

\section{Evaluation Metrics and Reasoning Criteria}
\label{sec.evaluation_metrics}
To evaluate the performance of SilVar-Med, we used both traditional text-based evaluation metrics and a novel LLM-as-Judge assessment. Traditional metrics include BLEU, ROUGE, and BERTScore, which measure the textual similarity between the model’s generated responses and ground truth references. However, these methods may not fully capture the accuracy and reasoning quality of medical abnormality detection.

To address this limitation, we propose an LLM-as-Judge evaluation framework to evaluate the reasoning of SilVar-Med’s performance in medical domain. To make the justification clear and consistent, we define two key criteria: (1) the accuracy of abnormality predictions and (2) the reasoning behind each prediction. Here, we measure two factors which are the structure of the answer and the accuracy of the answer. In terms of accuracy, the framework categorizes model responses into four levels:

\begin{itemize}
\renewcommand{\labelitemi}{}
\setlength{\itemindent}{-0.3cm} 
    \item \textbf{0: Completely Incorrect} – The prediction fails to answer the question, is off-topic, or entirely unrelated to the ground truth.
    \item \textbf{1: Significantly Incorrect} – The prediction attempts to answer the question but does not match the ground truth in terms of understanding, terminology, or core explanation.
    \item \textbf{2: Partially Correct} – The prediction directly answers the question and provides an explanation. Both the answer and the explanation reflect a reasonable understanding of the main idea, though they contain minor irrelevant or incorrect information.
    \item \textbf{3: Fully Correct} – The prediction completely aligns with the ground truth, providing both a clear answer and a well-reasoned explanation.
\end{itemize}

By adopting this approach, we move beyond a strict right/wrong classification and enable medical professionals to interpret model outputs, particularly in cases where the model exhibits uncertainty. To implement this evaluation, we use several commercial large language models including GPT-4o and Gemini Flash 1.5, to assess the responses. We then compute Pearson Correlation and Spearman Correlation to analyze the consistency between the LLM-based assessments and traditional metrics.

Beyond automated evaluations, three medical imaging specialists from our team independently assess SilVar-Med’s predictions. We then compare their evaluations with the results obtained from GPT-4o and Gemini, ensuring a comprehensive assessment that combines both expert judgment and automated analysis. 

\section{Experimental Result}

\subsection{Speech-To-Text Quality}

Before integrating Whisper to SilVar-Med, we fine-tuned it using a combination of the VQA-RAD, English SLAKE, and VQA-Med 2019 datasets, as outlined in \ref{sec.train_pipeline}. It is important to note that we convert the questions of these datasets to speech to train the model because we want to maximize the performance of audio encoder in the medical domain. Here, we evaluate the performance of two Whisper models (Tiny and Small) using Word Error Rate (WER) and Character Error Rate (CER), which are standard benchmarks for speech-to-text accuracy \cite{chen1998evaluation, morris2004and}.

\begin{table}[h]
\centering
\small
\begin{tabular}{lcccc}
\toprule
\textbf{Models} & \multicolumn{2}{c}{\textbf{WER}} & \multicolumn{2}{c}{\textbf{CER}} \\ \cmidrule(lr){2-3}  \cmidrule(lr){4-5}
 & train & test & train & test \\ \midrule
Whisper Tiny & 2.01 & 2.67 & 2.01 & 2.99 \\
Whisper Small & 2.02 & 4.57 & 1.59 & 3.50 \\
\bottomrule
\end{tabular}
\caption{Evaluation of audio encoder baselines (Whisper Tiny and Whisper Small) using WER and CER on the combined dataset (VQA-RAD, English SLAKE, and VQA-Med 2019).}
\label{tab.result-whisper}
\end{table}

The results, presented in \ref{tab.result-whisper}, indicate that Whisper Tiny and Whisper Small achieve comparable performance, with variations across WER and CER metrics. Specifically, Whisper Tiny achieves a WER of 2.01\% (train) and 2.67\% (test), along with a CER of 2.01\% (train) and 2.99\% (test). Whisper Small, on the other hand, reports a WER of 2.02\% (train) and 4.57\% (test), with a CER of 1.59\% (train) and 3.50\% (test).

Interestingly, while Whisper Small attains a lower CER during training (1.59\% vs. 2.01\%), it exhibits a significantly higher WER on the test set (4.57\% vs. 2.67\%), suggesting that it may be more prone to overfitting compared to Whisper Tiny. This discrepancy indicates that while the Small model has better character-level accuracy in training, its generalization to unseen test data is weaker. Given this observation, the Whisper Tiny model appears to be the more stable choice, balancing both WER and CER more consistently across training and testing phases. Moreover, since Whisper Tiny has a smaller number of parameters compared to Whisper Small, it is computationally more efficient. This makes it a more practical choice for our end-to-end fine-tuning process, as it reduces training time and resource consumption while still maintaining strong performance. Furthermore, these results reinforce the feasibility of using Google Cloud APIs to generate synthetic voice data without considering emotional expressiveness, as the overall error rates remain relatively low.

\subsection{Speech-Driven Medical VLMs}


To evaluate SilVar-Med’s performance, we evaluated it on the test set using BERTScore, BLEU, and ROUGE as standard text generation metrics.
Since there are no established benchmarks for speech-driven VLMs in the medical domain and only a few existing speech-driven VLM models, we compared SilVar-Med’s performance against SilVar and commercial speech-driven vision-language models (VLMs), including GPT-4o Mini and Gemini Flash 1.5. Unlike SilVar-Med, which is an end-to-end speech-driven VLM, GPT-4o Mini and Gemini Flash 1.5 follow a cascaded approach, requiring an intermediate step to convert audio into text before processing.

\begin{table}[h]
    \centering
    \small
    \begin{tabular}{lccc}
        \toprule
        \textbf{Models} & \textbf{BertScore} & \textbf{BLEU} & \textbf{ROUGE} \\
        \midrule
        SilVar-Med (Llama 3.1)& 0.82 & 20.87 \% & 55.18 \% \\
        GPT-4o mini& 0.76 & 7.25 \% & 46.33 \% \\
        Gemini Flash 1.5 & 0.75 & 3.32 \% &  34.07 \% \\

        \addlinespace[3pt] 
        \midrule
        \multicolumn{4}{l}{\textit{Ablation study with different language models for SilVar-Med}} \\
        \addlinespace[3pt] 
        SilVar-Med (Deepseek)& 0.81 & 20.43 \% & 54.45 \% \\
        \bottomrule
    \end{tabular}
    \caption{Comparison between the SilVar-Med and speech-driven VLMs on the test set.}
    \label{tab:result_SilVar_test_speech}
\end{table}

The results, summarized in \ref{tab:result_SilVar_test_speech}, indicate that SilVar-Med consistently outperforms GPT-4o mini and Gemini Flash 1.5 across all evaluated metrics. With a BERTScore of 0.82, SilVar-Med demonstrates a stronger semantic alignment with ground truth responses compared to GPT-4o mini (0.76) and Gemini Flash 1.5 (0.75), reflecting its ability to generate contextually accurate medical explanations. In terms of BLEU, SilVar-Med achieves 20.87\%, significantly surpassing GPT-4o mini (7.25\%) and Gemini Flash 1.5 (3.32\%), indicating superior syntactic and lexical accuracy in structured medical reasoning. Additionally, SilVar-Med attains the highest ROUGE score of 55.18\%, outperforming GPT-4o mini (46.33\%) and Gemini Flash 1.5 (34.07\%). This suggests that SilVar-Med more effectively captures key phrases and maintains coherence with reference texts.

Overall, these findings indicate that SilVar-Med’s domain-specific fine-tuning enables it to generate clinically relevant and semantically precise explanations, making it highly suitable for medical VQA tasks with end-to-end speech queries.



\subsection{Reasoning Ability and Human Evaluation}

\begin{table}[h]
    \centering
    \small
    \begin{tabular}{lcccc}
        \toprule
        \textbf{GPT4o} & \textbf{Gemini} & \textbf{Exp. 1} & \textbf{Exp. 2} & \textbf{Exp. 3} \\
        \midrule
        143/148 & 145/148 & 146/148 & 145/148 & 145/148 \\
        \bottomrule
    \end{tabular}
    \caption{Evaluation of the prediction structure of SilVar-Med models using GPT-4o mini, Gemini 1.5 Flash, and human experts  (denoted as Exp. in the table). }
    \label{tab:result_structure}
\end{table}

To evaluate SilVar-Med’s reasoning capabilities, we evaluate its response structure and reasoning accuracy of predictions using the LLM-as-Judge framework (GPT-4o and Gemini Flash 1.5) together with expert evaluations. For \textbf{structural responses}, we first analyze whether SilVar-Med’s responses follow a coherent and structured format, as this is essential for medical interpretability. The results is shown in \ref{tab:result_structure}, in which, the scores of GPT-4o mini and Gemini Flash 1.5 are 143/148 and 145/148, respectively. In addition, our expert evaluations further reinforce these findings, with scores reaching 146/148, 145/148, and 145/148, indicating that the model generally maintains a structured response format that aligns with human expectations.

\begin{table}[h]
    \centering
    \small
    \begin{tabular}{lccccc} 
        \toprule
        \begin{tabular}{@{}l@{}}\textbf{Reasoning}\\\textbf{accuracy}\end{tabular}  & \textbf{Exp 1} & \textbf{Exp 2} & \textbf{Exp 3} & \textbf{GPT4o} & \textbf{Gemini} \\
        \midrule
        \addlinespace[3pt] 
        \rowcolor{gray!10}\multicolumn{6}{l}{\textit{SilVar-Med with the langue module of Llama 3.1 8B}}\\
        \addlinespace[5pt] 
        \begin{tabular}{@{}l@{}}Completely\\Incorrect\end{tabular}  & 11 & 6 & 13 & 39.00 & 22.00 \\
        \begin{tabular}{@{}l@{}}Significantly\\Incorrect\end{tabular}  & 28 & 30 & 33 & 9.67 & 23.67 \\
        \begin{tabular}{@{}l@{}}Partially\\Correct\end{tabular} & 13 & 15 & 28 & 39.67 & 54.00 \\
        \begin{tabular}{@{}l@{}}\textbf{Fully}\\\textbf{Correct}\end{tabular} & 96 & 97 & 74 & 59.67 & 48.33 \\

        \midrule
        \addlinespace[3pt] 
        \rowcolor{gray!10}\multicolumn{6}{l}{\textit{Ablation studies of SilVar-Med with the langue module of }}\\ 
        \rowcolor{gray!10}\multicolumn{6}{l}{\textit{Deepseek R1 Distill 8B}} \\
        \addlinespace[3pt] 
        \begin{tabular}{@{}l@{}}Completely\\Incorrect\end{tabular}  & 12 & 10 & 10 & 40.00 & 20.67\\
        \begin{tabular}{@{}l@{}}Significantly\\Incorrect\end{tabular}  & 39 & 41 & 47 & 8.67  & 23.00\\
        \begin{tabular}{@{}l@{}}Partially\\Correct\end{tabular} & 13 & 11 & 21 & 41.00  & 52.67\\
        \begin{tabular}{@{}l@{}}\textbf{Fully}\\\textbf{Correct}\end{tabular} & 84 & 86 & 70 & 58.33  & 51.67\\

        \bottomrule
    \end{tabular}
    \caption{Assessment of SilVar-Med’s reasoning accuracy behind abnormality prediction. The table compares expert evaluations (Exp. 1–3) with LLM-as-Judge assessments (GPT-4o and Gemini Flash 1.5). It is important to note that, Fully Correct denotes predictions that are both accurate and well-explained.}
    \label{tab:result_reasoning}
\end{table}

To assess the \textbf{reasoning accuracy} of SilVar-Med, we evaluate how well the model provides observations and justifications for its predictions. The reasoning accuracy is categorized into four levels: Completely Incorrect, Significantly Incorrect, Partially Correct, and Fully Correct, as shown in \ref{tab:result_reasoning}. A model is considered capable of reasoning-based abnormality detection if it can accurately respond to speech-driven medical queries while providing a coherent and justifiable explanation. Given the inherent variability in text generation by GPT-4o mini and Gemini Flash 1.5, we conducted three independent evaluation rounds per model and averaged the results to ensure consistency. In addition, three experts independently assessed the model outputs to provide a human benchmark for comparison.

\ref{tab:result_reasoning} indicates notable discrepancies between expert evaluations and LLM-based assessments. Experts rate more responses as Fully Correct (74–97 for Llama 3.1 and 70–86 for Deepseek R1 Distill) compared to GPT-4o (59.67–58.33) and Gemini (48.33–51.67). Gemini is more conservative, labeling a higher number of responses as Partially Correct, while GPT-4o assigns more Completely Incorrect ratings. Overall, \ref{tab:result_reasoning} shows that SilVar-Med demonstrates strong reasoning accuracy, effectively answering speech-driven medical queries with high prediction accuracy and well-structured explanations.

Despite the self-corrected and distilled learning in the general domain of Deepseek R1 8B Distill, we found that it achieves modest performance when integrated into SilVar-Med for medical abnormality detection. Additionally, the inconsistencies between GPT-4o, Gemini, and expert evaluations highlight the limitations of the LLM-as-Judge framework. While automated assessments provide useful insights, expert evaluation remains essential to ensure a balanced and clinically relevant assessment. 

\begin{table*}[h]
    \centering
    \small
    \begin{tabular}{lccccccc}
        \toprule
        \textbf{Models} & \textbf{Instruction} & \multicolumn{3}{c}{\textbf{SLAKE}} & \multicolumn{3}{c}{\textbf{VQA-RAD}}  \\
        \cmidrule(lr){3-5} \cmidrule(lr){6-8} 
        &  & \textbf{Ref} & \textbf{Open} & \textbf{Closed} & \textbf{Ref} & \textbf{Open} & \textbf{Closed}  \\
        \midrule
        \rowcolor{gray!10}\multicolumn{8}{l}{\textit{Representatives of existing studies in the literature}}\\
        \addlinespace[5pt] 
        LLaVA  \cite{NEURIPS2023_6dcf277e} & Text &  & 78.18 & 63.22 &  & 50.0 & 65.07    \\
        LLaVA-Med (From LLaVA) \cite{li2024llava}  & Text &  & 83.08 & 85.34 &  & 61.52 & 84.19   \\
        LLaVA-Med (BioMed CLIP) \cite{li2024llava}  & Text &  & 87.11 & 86.78 &  & 64.75 & 83.09   \\
        LLaVA-Med++ (w/ Medtrinity) \cite{xie2024medtrinity} &  Text &  & 86.20 & 89.20 &  & 77.10 & 86.00   \\
        LLaVA-Med++ (w/o Medtrinity) \cite{xie2024medtrinity} &  Text &  & 79.30 & 84.00 &  & 64.60 & 77.00   \\     
        
        
        MMBERT General \cite{khare2021mmbert} &  Text &  & - & - &  & 63.10 & 77.90   \\
        MEVF+SAN \cite{nguyen2019overcoming} &  Text &  & - & - &  & 40.70 & 74.10   \\
        CR \cite{zhan2020medical} &  Text &  & - & - &  & 60.00 & 79.30   \\
        Q2ATransformer \cite{liu2023q2atransformer} & Text  &  &  &  & 79.19 &  & 81.20    \\
        PubMedCLIP \cite{eslami2023pubmedclip} &  Text & 78.40 &  & 82.50 & 60.10 &  & 80.00    \\
        BiomedCLIP \cite{zhang2023biomedclip} &  Text & 82.05 &  & 89.7 & 67.60 &  & 79.80    \\
        M2I2 \cite{li2023self} & Text  & 74.70 &  & 91.10 & 66.50 &  & 83.50    \\
        
        \midrule
        \addlinespace[3pt]
        \rowcolor{gray!10}\multicolumn{8}{l}{\textit{SilVar-based studies with our own experiment}}\\
        \addlinespace[5pt]
        SilVar-Med 3.1 8B (Llama 3.1-8B)  & Speech &  & 74.08 & 79.44  &  & 55.34 & 62.56   \\
        SilVar-Med 3.1 8B (Llama 3.1-8B)  & Text &  & 74.32 & 80.03&  & 55.21 & 60.86   \\

        \addlinespace[3pt]
        \rowcolor{gray!10}\multicolumn{8}{l}{\textit{Ablation studies of SilVar-Med using different language models for the decoder}}\\
        \addlinespace[5pt]
        SilVar-Med DR8B (Deepseek R1 Distill-Llama-8B)  & Speech &  & 76.50 & 83.80&  & 58.85 & 68.35   \\
        SilVar-Med DR8B (Deepseek R1 Distill-Llama-8B)  & Text &  & 77.12 & 82.11&  & 60.31 & 67.98   \\
        SilVar-Med 2 7B (Llama 2)  & Speech &  & 73.23  & 76.34 &  &  54.75  & 57.77   \\
        SilVar-Med 2 7B (Llama 2)  & Text &  & 64.21 & 75.54&  & 55.65 &  75.78   \\

        \bottomrule
    \end{tabular}
    \caption{Comparison of SilVar-Med with various text-based medical VLMs on the SLAKE and VQA-RAD datasets. Results are reported for both open-ended and closed-ended questions, with reference-based scores where applicable. LLaVA-based and other state-of-the-art (SoTA) models rely on text input, while SilVar-Med processes speech-driven queries.}
    \label{tab:merged_results}
\end{table*}

\subsection{Evaluation on image-text VLMs Benchmarks}

While SilVar-Med is a speech-driven medical VLM, no medical speech-driven VLMs currently exist for direct comparison. Our objective in this evaluation is not to achieve SOTA performance but rather to demonstrate the potential of voice-based medical communication with VLMs. To provide context for SilVar-Med’s performance, we compare it with existing text-based medical VLMs across multiple datasets, including SLAKE, VQA-RAD, and Medical VQA 2019. Although our primary focus is on developing a speech-driven instruction-based medical VLM, we also include comparisons with its text-based counterparts. The evaluation results are presented in \ref{tab:merged_results} and \ref{tab:result_VQA-2019}.

\begin{table}[h]
    \centering
    \small

    \begin{tabular}{lccc}
        \toprule
        \textbf{Models}  & \textbf{Accuracy} &  \textbf{BLEU} & \textbf{Bert-sim} \\
        \midrule
        \rowcolor{gray!10}\multicolumn{4}{l}{\textit{Representatives of existing studies in the literature}}\\
        ImageCLEF \cite{ImageCLEFVQA-Med2019}    &  62.4 \%  &  64.4 \% & - \\
        MedVINT \cite{zhang2023pmc} &      &   & 0.63 \\
        Med-Flamingo \cite{moor2023med}    &   &   & 0.65 \\
        MMBERT \cite{khare2021mmbert} & 67.2 \%  & 69.0 \%  & \\
        CGMVQA \cite{ren2020cgmvqa}&  60.0 \% & 61.9 \%  &     \\
        VGG16+BERT \cite{yan2019zhejiang} & 62.4 \% & 64.4 \% &  \\
        \midrule
        \addlinespace[3pt]
        \rowcolor{gray!10}\multicolumn{4}{l}{\textit{SilVar-based studies with our own experiment}}\\
        \addlinespace[5pt]
        SilVar-Med 3.1 8B (speech) &  64.99\% &  62.24 \% & 0.80 \\
        SilVar-Med 3.1 8B (text) & 66.10\% & 63.11\%& 0.80\\
        \addlinespace[3pt]
        \rowcolor{gray!10}\multicolumn{4}{l}{\textit{Ablation studies}}\\
        \addlinespace[5pt]
        SilVar-Med DR8B (speech) & 68.03\% & 69.12\%& 0.80\\
        SilVar-Med DR8B (text) & 68.77\% & 68.34\%& 0.81\\
        SilVar-Med 2 7B  (speech) & 66.36\% & 61.93\% & 0.79\\
        SilVar-Med 2 7B  (text) & 67.02\% & 59.17\% & 0.80\\
        \bottomrule
    \end{tabular}    
    \caption{Evaluation results of SilVar-Med with different text-based medical VLMs on the Medical VQA 2019 dataset.}
    \label{tab:result_VQA-2019}
\end{table}

\begin{table*}[ht]
\centering
\small
\begin{tabular}{>{\arraybackslash}m{12cm} >{\centering\arraybackslash}m{4.5cm}}
\toprule
\rowcolor{gray!10} \multicolumn{2}{l}{\textbf{Example 1}: Abnormality detection.} \\

\includegraphics[height=10pt]{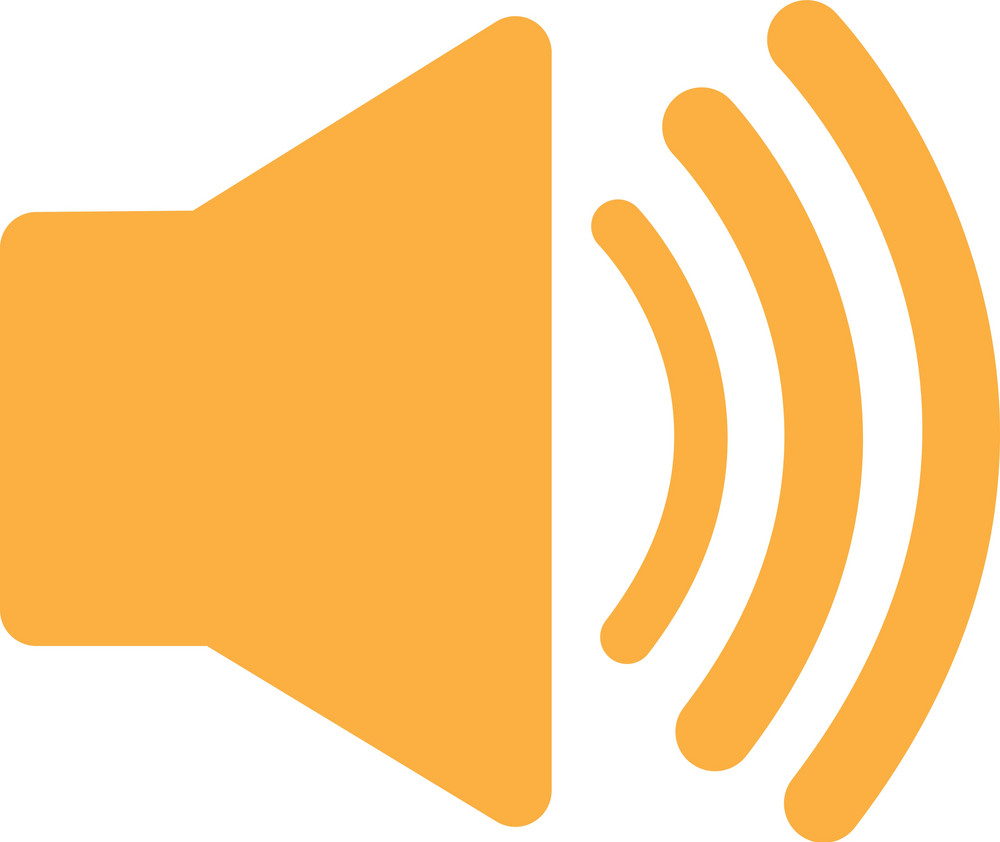} \hspace{0.1cm} Does the lung look abnormal?\\ \vspace{-0.5cm}\hspace{4.4cm} \includegraphics[height=80pt]{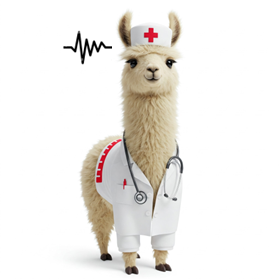}

\hspace{-0.45cm}\textbf{Ground truth}: Yes, \textcolor{purple}{the lung appears abnormal as there are irregularities} in the lung fields that may indicate potential pathology. The presence of unusual opacities or changes in the lung structure can suggest conditions such as \textcolor{teal}{infection, inflammation, or other lung diseases}. &  \includegraphics[width=3.5cm]{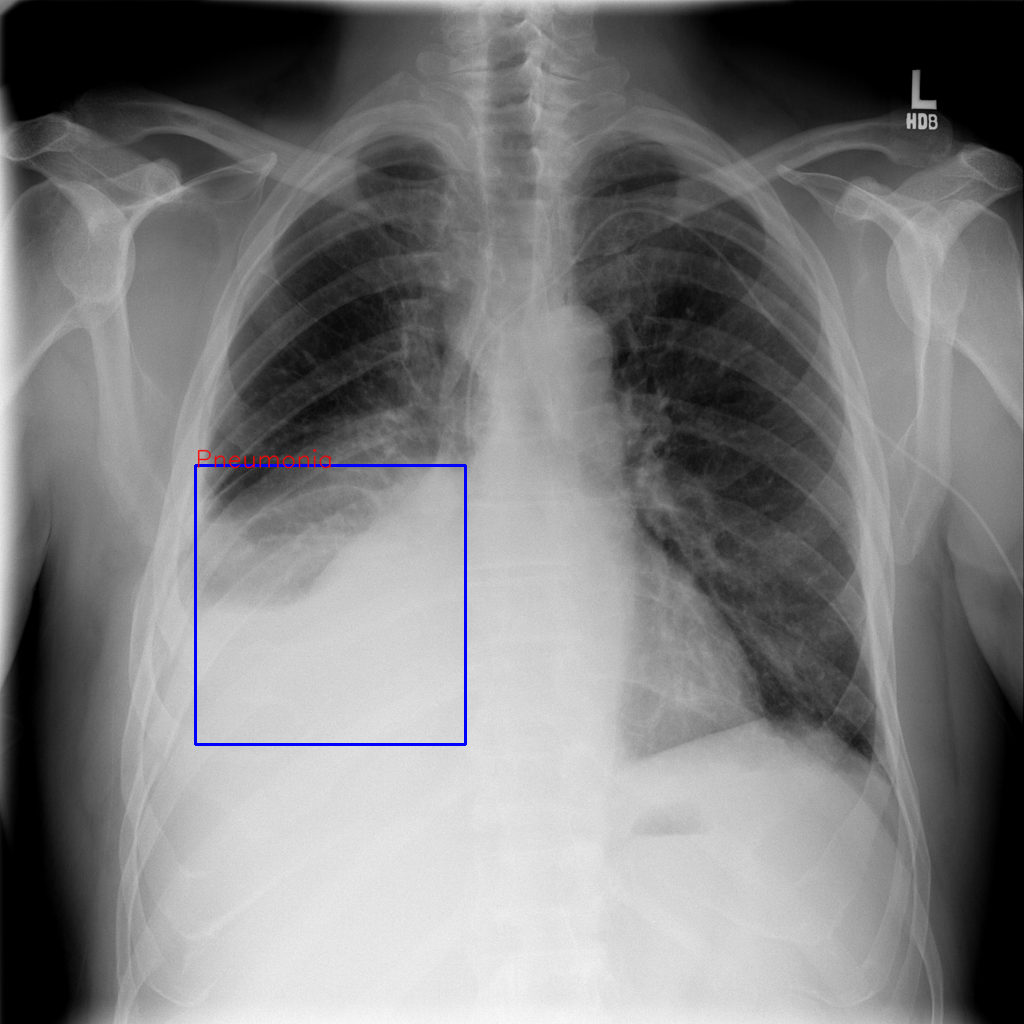} \\ \midrule
\multicolumn{2}{p{17cm}}{\textbf{SilVar-Med (Llama 3.1)}: Yes, \textcolor{purple}{the lung appears abnormal due to the presence of irregularities} in the lung fields, which may indicate potential pathology such as \textcolor{teal}{infection, inflammation, or other lung conditions}. The overall lung structure shows signs of asymmetry or density changes that deviate from normal anatomy. These findings warrant further investigation to determine the underlying cause.}\\ \midrule
\multicolumn{2}{p{17cm}}{\textbf{GPT-4o mini}: \textcolor{purple}{The lung appears to be within normal} limits based on the provided X-ray image. There are \textcolor{teal}{no obvious signs of abnormalities such as masses, infiltrates, or significant opacities}. The lung fields are clear, and the vascular markings are normal, indicating healthy lung tissue. However, a definitive assessment should be made by a qualified radiologist for a comprehensive evaluation.}\\ \midrule
\multicolumn{2}{p{17cm}}{\textbf{Gemini 1.5 Flash}: Based on the provided chest X-ray, \textcolor{purple}{the lungs appear normal}.  There is \textcolor{teal}{no visible evidence of infiltrates, masses, or pneumothorax}. The lung fields demonstrate normal aeration and the heart size appears within normal limits.  Further evaluation may be necessary depending on the clinical presentation.}\\ 
\bottomrule
\end{tabular}
\caption{Comparison of prediction between our models and the other speech-driven model on the reasoning abnormal detection. Unlike GPT-4o and Gemini 1.5 Flash, our SilVar-Med is an end-to-end speech-driven VLM. For more demonstration, please visit \href{https://github.com/Hanhpt23/SilVar}{SilVar-Med}.}
\label{tab:chatbot_comparison}
\end{table*}

\textbf{Performance on SLAKE}: SilVar-Med (Llama 3.1-8B, speech-based) achieves an accuracy of 74.08\% on SLAKE (Open QA) and 79.44\% on SLAKE (Closed QA). Compared to LLaVA-Med++ (Medtrinity), which achieves 86.2\% (Open) and 89.2\% (Closed), SilVar-Med still has room for improvement, particularly in open-ended responses. However, the gap is smaller when comparing against LLaVA-Med, where SilVar-Med’s performance remains competitive. It is important to note that most of the models in \ref{tab:merged_results} are text based, and not able to generate reasoning behind prediction. Additionally, we also use text as direct input for language models. As a result, there is a small performance gap when using text-based input versus audio-based input for SilVar-Med.

\textbf{Performance on VQA-RAD}: Similarly, SilVar-Med achieves an accuracy of 55.34\% on VQA-RAD (Open QA) and 62.56\% on VQA-RAD (Closed QA). Compared to LLaVA-Med++ (Medtrinity), which achieves 77.1\% (Open) and 86.0\% (Closed), SilVar-Med exhibits lower performance, particularly in open-ended responses. However, when compared to earlier LLaVA-Med models, such as LLaVA-Med (BioMed CLIP) with 64.75\% (Open) and 83.09\% (Closed), the performance gap is narrower. Notably, SilVar-Med's performance surpasses several traditional VLMs, such as MEVF+SAN and is competitive with models like CR.

\textbf{Performance on Medical VQA 2019}: For the Medical VQA 2019 dataset (\ref{tab:result_VQA-2019}), SilVar-Med achieves an accuracy of 64.99\%, outperforming models like ImageCLEF (62.4\%) and VGG16+BERT (62.4\%), while being competitive with MMBERT (67.2\%). In terms of BLEU score (62.24), SilVar-Med performs comparably to other models, indicating strong textual coherence. The BERT similarity score (0.80) is higher than MedVINT (0.63) and Med-Flamingo (0.65), suggesting that SilVar-Med’s responses are more semantically aligned with the ground truth. These results demonstrate that SilVar-Med’s performance is strong among speech-based models and is comparable to leading text-based models. The inclusion of a speech interface provides additional usability advantages in medical applications where hands-free interactions are crucial.

\section{Ablation Study}
We conducted experiments with the language models in SilVar-Med by using different models, including Llama 2 and DeepSeek R1 (Distill-Llama-8B), to analyze their impact on SilVar-Med’s performance. By testing SilVar-Med with multiple LLMs, we aim to identify the optimal configuration for medical abnormality detection and reasoning tasks. We evaluate the model’s performance on both our dataset and published datasets, as shown in \ref{tab:result_reasoning}, \ref{tab:merged_results}, and \ref{tab:result_VQA-2019}. 

In \ref{tab:result_reasoning}, there is a small discrepancy between the performance of SilVar-Meds when using Llama 3.1 and DeepSeek R1 (Distill-8B), despite DeepSeek’s distilled knowledge and reasoning ability in the general domain. In addition, the performance of SilVar-Med with Llama 2 is competitive compared to DeepSeek and Llama 3.1 on the Slake, VQA-RAD, and VQA 2019 datasets, as shown in \ref{tab:merged_results} and \ref{tab:result_VQA-2019}. Compared to direct text input, the audio-driven approach achieves comparable performance or performs on par, despite challenges related to speech conversion errors, variations in spoken queries, and audio embeddings. These results highlight the robustness and adaptability of SilVar-Med, demonstrating its effectiveness across different language models and datasets.

\begin{table}[h]
    \centering
    \small
    \begin{tabular}{lcccc}
        \toprule
        \textbf{Prompts} & \multicolumn{2}{c}{\textbf{Reasoning Accuracy}} & \textbf{BERTScore}  & \textbf{BLEU} \\ \cmidrule(lr){2-3}  
        & GPT4o & Gemini &  &  \\
 
        \midrule
        Zero-shot &58.33 & 51.67& 0.80 & 21.43\% \\
        COT  & 61 & 50 & 0.81 & 22.16\% \\
        TOT  & 59 & 47 & 0.80 & 21.44\% \\
        \bottomrule
    \end{tabular}
    \label{tab.ablationstudy}
    \caption{Comparison of SilVar-Med using different prompts.}
\end{table}

Furthermore, to investigate the reasoning ability of SilVar-Med in medical reasoning tasks, we conduct an ablation study by employing Chain-of-Thought (CoT) and Tree-of-Thought (ToT) prompting techniques. As shown in Tab. 10, we use GPT-4o and Gemini Flash 1.5 to evaluate the model's performance. The results indicate that structured reasoning techniques such as CoT and ToT might improve the model’s performance compared to zero-shot prompting, although not significantly in our study. 

\section{Conclusion}

In this study, we demonstrate a proof-of-concept study for speech-driven medical VLMs, focusing on reasoning for abnormality detection and interpretable AI assessments. We address two key challenges: (1) enabling voice communication in medical VLMs and (2) providing reasoning for each abnormality prediction. To this end, we also introduce a reasoning dataset for training and testing. The result is evaluated by three physicians along with a proposed LLM-as-Judge evaluation framework to assess both the accuracy and reasoning quality of its predictions. 

Our experiments with reasoning interpretation, demonstrate the effectiveness of SilVar-Med in generating structured, accurate, and interpretable medical responses. Despite the challenge of speech-driven input, the model performs on par with other models. In terms of reasoning, although our work is limited by the dataset and the MRI, CT, and X-ray modalities, it provides reliable reasoning and demonstrates its potential in the medical domain, addressing the weaknesses of SOTA models. We also demonstrate that by minimizing speech-to-text errors, the model yields high-quality audio embeddings, leading to performance comparable to text-based models. Additionally, we found a lack of available speech-driven datasets benchmark for medical VLMs, highlighting a critical gap in the advancing field.

\balance
{\small
\bibliographystyle{ieee_fullname}
\bibliography{egbib}

\begin{thebibliography}{10}\itemsep=-1pt

\bibitem{achiam2023gpt}
Josh Achiam, Steven Adler, Sandhini Agarwal, Lama Ahmad, Ilge Akkaya, Florencia~Leoni Aleman, Diogo Almeida, Janko Altenschmidt, Sam Altman, Shyamal Anadkat, et~al.
\newblock Gpt-4 technical report.
\newblock {\em arXiv preprint arXiv:2303.08774}, 2023.

\bibitem{alayrac2022flamingo}
Jean-Baptiste Alayrac, Jeff Donahue, Pauline Luc, Antoine Miech, Iain Barr, Yana Hasson, Karel Lenc, Arthur Mensch, Katherine Millican, Malcolm Reynolds, et~al.
\newblock Flamingo: a visual language model for few-shot learning.
\newblock {\em Advances in neural information processing systems}, 35:23716--23736, 2022.

\bibitem{alkhaldi2024minigpt}
Asma Alkhaldi, Raneem Alnajim, Layan Alabdullatef, Rawan Alyahya, Jun Chen, Deyao Zhu, Ahmed Alsinan, and Mohamed Elhoseiny.
\newblock Minigpt-med: Large language model as a general interface for radiology diagnosis.
\newblock {\em arXiv preprint arXiv:2407.04106}, 2024.

\bibitem{awadalla2023openflamingo}
Anas Awadalla, Irena Gao, Josh Gardner, Jack Hessel, Yusuf Hanafy, Wanrong Zhu, Kalyani Marathe, Yonatan Bitton, Samir Gadre, Shiori Sagawa, et~al.
\newblock Openflamingo: An open-source framework for training large autoregressive vision-language models.
\newblock {\em arXiv preprint arXiv:2308.01390}, 2023.

\bibitem{bae2023ehrxqa}
Seongsu Bae, Daeun Kyung, Jaehee Ryu, Eunbyeol Cho, Gyubok Lee, Sunjun Kweon, Jungwoo Oh, Lei Ji, Eric Chang, Tackeun Kim, et~al.
\newblock Ehrxqa: A multi-modal question answering dataset for electronic health records with chest x-ray images.
\newblock {\em Advances in Neural Information Processing Systems}, 36:3867--3880, 2023.

\bibitem{bai2023qwen}
Jinze Bai, Shuai Bai, Yunfei Chu, Zeyu Cui, Kai Dang, Xiaodong Deng, Yang Fan, Wenbin Ge, Yu Han, Fei Huang, et~al.
\newblock Qwen technical report.
\newblock {\em arXiv preprint arXiv:2309.16609}, 2023.

\bibitem{ImageCLEFVQA-Med2019}
Asma {Ben Abacha}, Sadid~A. Hasan, Vivek~V. Datla, Joey Liu, Dina Demner-Fushman, and Henning M\"uller.
\newblock Vqa-med: Overview of the medical visual question answering task at imageclef 2019.
\newblock In {\em Working Notes of {CLEF} 2019}, volume 2380 of {\em {CEUR} Workshop Proceedings}, Lugano, Switzerland, September 9-12 2019. CEUR-WS.org.

\bibitem{blankemeier2024merlin}
Louis Blankemeier, Joseph~Paul Cohen, Ashwin Kumar, Dave Van~Veen, Syed Jamal~Safdar Gardezi, Magdalini Paschali, Zhihong Chen, Jean-Benoit Delbrouck, Eduardo Reis, Cesar Truyts, et~al.
\newblock Merlin: A vision language foundation model for 3d computed tomography.
\newblock {\em Research Square}, pages rs--3, 2024.

\bibitem{brown2020language}
Tom~B Brown.
\newblock Language models are few-shot learners.
\newblock {\em arXiv preprint arXiv:2005.14165}, 2020.

\bibitem{chen2023minigpt}
Jun Chen, Deyao Zhu, Xiaoqian Shen, Xiang Li, Zechun Liu, Pengchuan Zhang, Raghuraman Krishnamoorthi, Vikas Chandra, Yunyang Xiong, and Mohamed Elhoseiny.
\newblock Minigpt-v2: large language model as a unified interface for vision-language multi-task learning.
\newblock {\em arXiv preprint arXiv:2310.09478}, 2023.

\bibitem{chen2024medblip}
Qiuhui Chen and Yi Hong.
\newblock Medblip: Bootstrapping language-image pre-training from 3d medical images and texts.
\newblock In {\em Proceedings of the Asian Conference on Computer Vision}, pages 2404--2420, 2024.

\bibitem{chen1998evaluation}
Stanley~F Chen, Douglas Beeferman, and Roni Rosenfeld.
\newblock Evaluation metrics for language models.
\newblock 1998.

\bibitem{chen2024internvl}
Zhe Chen, Jiannan Wu, Wenhai Wang, Weijie Su, Guo Chen, Sen Xing, Muyan Zhong, Qinglong Zhang, Xizhou Zhu, Lewei Lu, et~al.
\newblock Internvl: Scaling up vision foundation models and aligning for generic visual-linguistic tasks.
\newblock In {\em Proceedings of the IEEE/CVF Conference on Computer Vision and Pattern Recognition}, pages 24185--24198, 2024.

\bibitem{deepseekR1}
DeepSeek-AI, Daya Guo, Dejian Yang, Haowei Zhang, Junxiao Song, Ruoyu Zhang, et~al.
\newblock Deepseek-r1: Incentivizing reasoning capability in llms via reinforcement learning, 2025.

\bibitem{dubey2024Llama}
Abhimanyu Dubey, Abhinav Jauhri, Abhinav Pandey, Abhishek Kadian, Ahmad Al-Dahle, Aiesha Letman, Akhil Mathur, Alan Schelten, Amy Yang, Angela Fan, et~al.
\newblock The llama 3 herd of models.
\newblock {\em arXiv preprint arXiv:2407.21783}, 2024.

\bibitem{eslami2023pubmedclip}
Sedigheh Eslami, Christoph Meinel, and Gerard De~Melo.
\newblock Pubmedclip: How much does clip benefit visual question answering in the medical domain?
\newblock In {\em Findings of the Association for Computational Linguistics: EACL 2023}, pages 1151--1163, 2023.

\bibitem{he2020pathvqa}
Xuehai He, Yichen Zhang, Luntian Mou, Eric Xing, and Pengtao Xie.
\newblock Pathvqa: 30000+ questions for medical visual question answering.
\newblock {\em arXiv preprint arXiv:2003.10286}, 2020.

\bibitem{hu2024omnimedvqa}
Yutao Hu, Tianbin Li, Quanfeng Lu, Wenqi Shao, Junjun He, Yu Qiao, and Ping Luo.
\newblock Omnimedvqa: A new large-scale comprehensive evaluation benchmark for medical lvlm.
\newblock In {\em Proceedings of the IEEE/CVF Conference on Computer Vision and Pattern Recognition}, pages 22170--22183, 2024.

\bibitem{hyland2023maira}
Stephanie~L Hyland, Shruthi Bannur, Kenza Bouzid, Daniel~C Castro, Mercy Ranjit, Anton Schwaighofer, Fernando P{\'e}rez-Garc{\'\i}a, Valentina Salvatelli, Shaury Srivastav, Anja Thieme, et~al.
\newblock Maira-1: A specialised large multimodal model for radiology report generation.
\newblock {\em arXiv preprint arXiv:2311.13668}, 2023.

\bibitem{jiang2023mistral}
Albert~Q Jiang, Alexandre Sablayrolles, Arthur Mensch, Chris Bamford, Devendra~Singh Chaplot, Diego de~las Casas, Florian Bressand, Gianna Lengyel, Guillaume Lample, Lucile Saulnier, et~al.
\newblock Mistral 7b.
\newblock {\em arXiv preprint arXiv:2310.06825}, 2023.

\bibitem{jin2019pubmedqa}
Qiao Jin, Bhuwan Dhingra, Zhengping Liu, William~W Cohen, and Xinghua Lu.
\newblock Pubmedqa: A dataset for biomedical research question answering.
\newblock {\em arXiv preprint arXiv:1909.06146}, 2019.

\bibitem{khare2021mmbert}
Yash Khare, Viraj Bagal, Minesh Mathew, Adithi Devi, U~Deva Priyakumar, and CV Jawahar.
\newblock Mmbert: Multimodal bert pretraining for improved medical vqa.
\newblock In {\em 2021 IEEE 18th International Symposium on Biomedical Imaging (ISBI)}, pages 1033--1036. IEEE, 2021.

\bibitem{lau2018dataset}
Jason~J Lau, Soumya Gayen, Asma Ben~Abacha, and Dina Demner-Fushman.
\newblock A dataset of clinically generated visual questions and answers about radiology images.
\newblock {\em Scientific data}, 5(1):1--10, 2018.

\bibitem{le2024litegpt}
Khai Le-Duc, Ryan Zhang, Ngoc~Son Nguyen, Tan-Hanh Pham, Anh Dao, Ba~Hung Ngo, Anh~Totti Nguyen, and Truong-Son Hy.
\newblock Litegpt: Large vision-language model for joint chest x-ray localization and classification task.
\newblock {\em arXiv preprint arXiv:2407.12064}, 2024.

\bibitem{li2024llava}
Chunyuan Li, Cliff Wong, Sheng Zhang, Naoto Usuyama, Haotian Liu, Jianwei Yang, Tristan Naumann, Hoifung Poon, and Jianfeng Gao.
\newblock Llava-med: Training a large language-and-vision assistant for biomedicine in one day.
\newblock {\em Advances in Neural Information Processing Systems}, 36, 2024.

\bibitem{li2023blip}
Junnan Li, Dongxu Li, Silvio Savarese, and Steven Hoi.
\newblock Blip-2: Bootstrapping language-image pre-training with frozen image encoders and large language models.
\newblock In {\em International conference on machine learning}, pages 19730--19742. PMLR, 2023.

\bibitem{li2023self}
Pengfei Li, Gang Liu, Lin Tan, Jinying Liao, and Shenjun Zhong.
\newblock Self-supervised vision-language pretraining for medial visual question answering.
\newblock In {\em 2023 IEEE 20th International Symposium on Biomedical Imaging (ISBI)}, pages 1--5. IEEE, 2023.

\bibitem{lin2023pmc}
Weixiong Lin, Ziheng Zhao, Xiaoman Zhang, Chaoyi Wu, Ya Zhang, Yanfeng Wang, and Weidi Xie.
\newblock Pmc-clip: Contrastive language-image pre-training using biomedical documents.
\newblock In {\em International Conference on Medical Image Computing and Computer-Assisted Intervention}, pages 525--536. Springer, 2023.

\bibitem{liu2021slake}
Bo Liu, Li-Ming Zhan, Li Xu, Lin Ma, Yan Yang, and Xiao-Ming Wu.
\newblock Slake: A semantically-labeled knowledge-enhanced dataset for medical visual question answering.
\newblock In {\em 2021 IEEE 18th International Symposium on Biomedical Imaging (ISBI)}, pages 1650--1654. IEEE, 2021.

\bibitem{liu2024improved}
Haotian Liu, Chunyuan Li, Yuheng Li, and Yong~Jae Lee.
\newblock Improved baselines with visual instruction tuning.
\newblock In {\em Proceedings of the IEEE/CVF Conference on Computer Vision and Pattern Recognition}, pages 26296--26306, 2024.

\bibitem{NEURIPS2023_6dcf277e}
Haotian Liu, Chunyuan Li, Qingyang Wu, and Yong~Jae Lee.
\newblock Visual instruction tuning.
\newblock In {\em Advances in Neural Information Processing Systems}, volume~36, pages 34892--34916. Curran Associates, Inc., 2023.

\bibitem{liu2023q2atransformer}
Yunyi Liu, Zhanyu Wang, Dong Xu, and Luping Zhou.
\newblock Q2atransformer: Improving medical vqa via an answer querying decoder.
\newblock In {\em International Conference on Information Processing in Medical Imaging}, pages 445--456. Springer, 2023.

\bibitem{moor2023med}
Michael Moor, Qian Huang, Shirley Wu, Michihiro Yasunaga, Yash Dalmia, Jure Leskovec, Cyril Zakka, Eduardo~Pontes Reis, and Pranav Rajpurkar.
\newblock Med-flamingo: a multimodal medical few-shot learner.
\newblock In {\em Machine Learning for Health (ML4H)}, pages 353--367. PMLR, 2023.

\bibitem{morris2004and}
Andrew~Cameron Morris, Viktoria Maier, and Phil~D Green.
\newblock From wer and ril to mer and wil: improved evaluation measures for connected speech recognition.
\newblock In {\em Interspeech}, pages 2765--2768, 2004.

\bibitem{nguyen2019overcoming}
Binh~D Nguyen, Thanh-Toan Do, Binh~X Nguyen, Tuong Do, Erman Tjiputra, and Quang~D Tran.
\newblock Overcoming data limitation in medical visual question answering.
\newblock In {\em Medical Image Computing and Computer Assisted Intervention--MICCAI 2019: 22nd International Conference, Shenzhen, China, October 13--17, 2019, Proceedings, Part IV 22}, pages 522--530. Springer, 2019.

\bibitem{openai2024gpt4}
OpenAI.
\newblock Gpt-4.
\newblock Available at \url{https://openai.com/gpt-4}, 2024.
\newblock Model used for dataset generation.

\bibitem{pham2024SilVar}
Tan-Hanh Pham, Hoang-Nam Le, Phu-Vinh Nguyen, Chris Ngo, and Truong-Son Hy.
\newblock Silvar: Speech driven multimodal model for reasoning visual question answering and object localization.
\newblock {\em arXiv preprint arXiv:2412.16771}, 2024.

\bibitem{radford2023robust}
Alec Radford, Jong~Wook Kim, Tao Xu, Greg Brockman, Christine McLeavey, and Ilya Sutskever.
\newblock Robust speech recognition via large-scale weak supervision.
\newblock In {\em International conference on machine learning}, pages 28492--28518. PMLR, 2023.

\bibitem{ranasinghe2023language}
Kanchana Ranasinghe and Michael~S Ryoo.
\newblock Language-based action concept spaces improve video self-supervised learning.
\newblock {\em Advances in Neural Information Processing Systems}, 36:74980--74994, 2023.

\bibitem{ren2020cgmvqa}
Fuji Ren and Yangyang Zhou.
\newblock Cgmvqa: A new classification and generative model for medical visual question answering.
\newblock {\em IEEE Access}, 8:50626--50636, 2020.

\bibitem{royer2024multimedeval}
Corentin Royer, Bjoern Menze, and Anjany Sekuboyina.
\newblock Multimedeval: A benchmark and a toolkit for evaluating medical vision-language models.
\newblock {\em arXiv preprint arXiv:2402.09262}, 2024.

\bibitem{singhal2022large}
Karan Singhal, Shekoofeh Azizi, Tao Tu, S~Sara Mahdavi, Jason Wei, Hyung~Won Chung, Nathan Scales, Ajay Tanwani, Heather Cole-Lewis, Stephen Pfohl, et~al.
\newblock Large language models encode clinical knowledge.
\newblock {\em arXiv preprint arXiv:2212.13138}, 2022.

\bibitem{team2023gemini}
Gemini Team, Rohan Anil, Sebastian Borgeaud, Yonghui Wu, Jean-Baptiste Alayrac, Jiahui Yu, Radu Soricut, Johan Schalkwyk, Andrew~M Dai, Anja Hauth, et~al.
\newblock Gemini: a family of highly capable multimodal models.
\newblock {\em arXiv preprint arXiv:2312.11805}, 2023.

\bibitem{touvron2023Llama}
Hugo Touvron, Thibaut Lavril, Gautier Izacard, Xavier Martinet, Marie-Anne Lachaux, Timoth{\'e}e Lacroix, Baptiste Rozi{\`e}re, Naman Goyal, Eric Hambro, Faisal Azhar, et~al.
\newblock Llama: Open and efficient foundation language models.
\newblock {\em arXiv preprint arXiv:2302.13971}, 2023.

\bibitem{touvron2023Llama2}
Hugo Touvron, Louis Martin, Kevin Stone, Peter Albert, Amjad Almahairi, Yasmine Babaei, Nikolay Bashlykov, Soumya Batra, Prajjwal Bhargava, Shruti Bhosale, et~al.
\newblock Llama 2: Open foundation and fine-tuned chat models.
\newblock {\em arXiv preprint arXiv:2307.09288}, 2023.

\bibitem{windsor2023visionlanguage}
Rhydian Windsor, Amir Jamaludin, Timor Kadir, and Andrew Zisserman.
\newblock Vision-language modelling for radiological imaging and reports in the low data regime.
\newblock In {\em Medical Imaging with Deep Learning}, 2023.

\bibitem{wu2023towards}
Chaoyi Wu, Xiaoman Zhang, Ya Zhang, Yanfeng Wang, and Weidi Xie.
\newblock Towards generalist foundation model for radiology.
\newblock {\em arXiv preprint arXiv:2308.02463}, 2023.

\bibitem{xie2024medtrinity}
Yunfei Xie, Ce Zhou, Lang Gao, Juncheng Wu, Xianhang Li, Hong-Yu Zhou, Sheng Liu, Lei Xing, James Zou, Cihang Xie, et~al.
\newblock Medtrinity-25m: A large-scale multimodal dataset with multigranular annotations for medicine.
\newblock {\em arXiv preprint arXiv:2408.02900}, 2024.

\bibitem{yan2019zhejiang}
Xin Yan, Lin Li, Chulin Xie, Jun Xiao, and Lin Gu.
\newblock Zhejiang university at imageclef 2019 visual question answering in the medical domain.
\newblock {\em CLEF (working notes)}, 85, 2019.

\bibitem{yang2024qwen2}
An Yang, Baosong Yang, Beichen Zhang, Binyuan Hui, Bo Zheng, Bowen Yu, Chengyuan Li, Dayiheng Liu, Fei Huang, Haoran Wei, et~al.
\newblock Qwen2. 5 technical report.
\newblock {\em arXiv preprint arXiv:2412.15115}, 2024.

\bibitem{you2023cxr}
Kihyun You, Jawook Gu, Jiyeon Ham, Beomhee Park, Jiho Kim, Eun~K Hong, Woonhyuk Baek, and Byungseok Roh.
\newblock Cxr-clip: Toward large scale chest x-ray language-image pre-training.
\newblock In {\em International Conference on Medical Image Computing and Computer-Assisted Intervention}, pages 101--111. Springer, 2023.

\bibitem{zhan2020medical}
Li-Ming Zhan, Bo Liu, Lu Fan, Jiaxin Chen, and Xiao-Ming Wu.
\newblock Medical visual question answering via conditional reasoning.
\newblock In {\em Proceedings of the 28th ACM International Conference on Multimedia}, pages 2345--2354, 2020.

\bibitem{zhang2023biomedgpt}
Kai Zhang, Jun Yu, Eashan Adhikarla, Rong Zhou, Zhiling Yan, Yixin Liu, Zhengliang Liu, Lifang He, Brian Davison, Xiang Li, et~al.
\newblock Biomedgpt: A unified and generalist biomedical generative pre-trained transformer for vision, language, and multimodal tasks.
\newblock {\em arXiv e-prints}, pages arXiv--2305, 2023.

\bibitem{zhang2023biomedclip}
Sheng Zhang, Yanbo Xu, Naoto Usuyama, Hanwen Xu, Jaspreet Bagga, Robert Tinn, Sam Preston, Rajesh Rao, Mu Wei, Naveen Valluri, et~al.
\newblock Biomedclip: a multimodal biomedical foundation model pretrained from fifteen million scientific image-text pairs.
\newblock {\em arXiv preprint arXiv:2303.00915}, 2023.

\bibitem{zhang2023pmc}
Xiaoman Zhang, Chaoyi Wu, Ziheng Zhao, Weixiong Lin, Ya Zhang, Yanfeng Wang, and Weidi Xie.
\newblock Pmc-vqa: Visual instruction tuning for medical visual question answering.
\newblock {\em arXiv preprint arXiv:2305.10415}, 2023.

\bibitem{zheng2023judging}
Lianmin Zheng, Wei-Lin Chiang, Ying Sheng, Siyuan Zhuang, Zhanghao Wu, Yonghao Zhuang, Zi Lin, Zhuohan Li, Dacheng Li, Eric.~P Xing, Hao Zhang, Joseph~E. Gonzalez, and Ion Stoica.
\newblock Judging llm-as-a-judge with mt-bench and chatbot arena, 2023.

\bibitem{zhu2023minigpt}
Deyao Zhu, Jun Chen, Xiaoqian Shen, Xiang Li, and Mohamed Elhoseiny.
\newblock Minigpt-4: Enhancing vision-language understanding with advanced large language models.
\newblock {\em arXiv preprint arXiv:2304.10592}, 2023.

\end{thebibliography}
}

\end{document}